\documentclass{article}

     \PassOptionsToPackage{numbers, compress}{natbib}
     
\usepackage{multirow}
\usepackage{tablefootnote}
\usepackage[preprint]{neurips_data_2023}

\usepackage{graphicx}
\usepackage[utf8]{inputenc} 
\usepackage[T1]{fontenc}    
\usepackage{hyperref}       
\usepackage{url}            
\usepackage{booktabs}       
\usepackage{amsfonts}       
\usepackage{nicefrac}       
\usepackage{microtype}      
\usepackage{xcolor}         

\title{3D-Speaker: A Large-Scale Multi-Device, Multi-Distance, and Multi-Dialect Corpus for Speech Representation Disentanglement}

\author{%
  Siqi Zheng, Luyao Cheng, Yafeng Chen, Hui Wang, Qian Chen \\
  DAMO Academy\\
  Alibaba Group\\
  \texttt{\{zsq174630, shuli.cly\}@alibaba-inc.com} \\
}

\begin{document}

\maketitle

\begin{abstract}
  Disentangling uncorrelated information in speech utterances is a crucial research topic within speech community. Different speech-related tasks focus on extracting distinct speech representations while minimizing the affects of other uncorrelated information. We present a large-scale speech corpus to facilitate the research of speech representation disentanglement. \textbf{3D-Speaker} contains over 10,000 speakers, each of whom are simultaneously recorded by multiple \textbf{D}evices, locating at different \textbf{D}istances, and some speakers are speaking multiple \textbf{D}ialects. The controlled combinations of multi-dimensional audio data yield a matrix of a diverse blend of speech representation entanglement, thereby motivating intriguing methods to untangle them. The multi-domain nature of 3D-Speaker also makes it a suitable resource to evaluate large universal speech models and experiment methods of out-of-domain learning and self-supervised learning. \url{https://3dspeaker.github.io/}
  
\end{abstract}

\section{Introduction}

  Disentangling uncorrelated information in speech utterances is a crucial research topic within speech community\cite{DBLP:conf/icassp/HsuZWCWWG19}\cite{DBLP:conf/nips/HsuZG17}. A speech utterance typically consists of a mixture of information such as content of speech, speaker characteristics, dialect, recording device, distance to the sound source, and other information such as environment and noise. Different speech-related tasks aim at recognizing the specific information of interest while minimizing the affects of uncorrelated information. For example, in automatic speech recognition (ASR), researchers aim at recognizing the content of speech without being affected by speakers' voice characteristics, noise, and other uncorrelated information. Speaker verification (SV), on the other hand, focuses on identifying speaker's voice, independent of the content of speech. In speech synthesis tasks, researchers leverage disentangled embeddings to achieve goals such as style transfer, cross-language synthesis, and voice conversion etc.

  Speaker verification is one of the tasks that benefit most from the successful disentanglement of different speech-related information, as speaker's voice is an omnipresent characteristics in every speech utterance, but is intricately mingled with other speech information, such as speech content, device, language, etc. It also possesses natures such as long-term stability and relative uniqueness. Methods and techniques to extract disentangled speaker representation from human speech can well be generalized to other machine learning fields, such as extracting global features in vision and natural language understanding. 
  
  However, research of speech representation disentanglement has largely been hindered by the lack of large-scale publicly-available dataset containing explicit labels characterizing multiple attributes of speech. In order to help accelerate the related research, we introduce 3D-Speaker, where all utterances contain labels depicting multiple speech characteristics, such as speaker ID, dialect spoken, type of recording device, and the distance from device to the speaker. 

  3D-Speaker can be used to experiment supervised and unsupervised methods, as well as in- and out-of-domain learning. It can also be used to evaluate universal speech models aiming to possess the ability to perform common speech-related tasks on any domain.
  
  According to previous studies, increasing the number of speakers in training data remarkably improves the performance of speaker verification system \cite{DBLP:conf/interspeech/ChungNZ18}\cite{DBLP:conf/interspeech/ZhengLSL19a}. To the best of our knowledge, 3D-Speaker is the largest publicly-accessible corpus in terms of number of speakers. 

\section{Related Works}

There are abundant efforts trying to extract speaker embeddings that represent only speakers' voice, removing impacts of uncorrelated information. These methods range from adversarial learning\cite{DBLP:conf/interspeech/BorgstromSRS17}\cite{DBLP:conf/interspeech/KatariaVZMD21}\cite{DBLP:conf/odyssey/TongZZXHL22}\cite{DBLP:conf/interspeech/ChenWQ20a}\cite{DBLP:conf/interspeech/ZhengLS20}, to data-driven approaches such as data augmentation and generalization\cite{DBLP:conf/aaai/LiYSH18}\cite{DBLP:conf/icassp/ShahnawazuddinA20}\cite{DBLP:conf/iconip/LiZLZYH22}\cite{DBLP:conf/icassp/ZhangWLLDC22}. Some speech representation models based on self-supervised are shown to have the ability to untangle different speech information into different layers\cite{DBLP:journals/jstsp/ChenWCWLCLKYXWZ22}\cite{DBLP:journals/taslp/HsuBTLSM21}\cite{DBLP:conf/nips/BaevskiZMA20}.

Several previously released corpus have successfully boosted research in speech recognition and speaker verification. VoxCeleb 1 \& 2 \cite{DBLP:conf/interspeech/NagraniCZ17} collected over 7000 speakers from the internet and the speakers span a wide range of different ethnicity groups, languages, and ages. Unfortunately, labels other than speaker identities are missing, making it less effective in disentangling other speech representations and tackling out-of-domain tasks.

CN-Celeb\cite{DBLP:conf/icassp/FanKLLCCZZCW20} collected around 3000 speakers in a way similar to VoxCeleb. Additionally, CN-Celeb provided the ``genre'' labels, which introduce more varieties into the corpus and potentially allows for ``cross-genre'' studies. However, a genre such as Play, Movie, Vlog, Drama, is not a direct speech characteristics and infers little about the disentangled speech representation of interest. 

The Librispeech\cite{DBLP:conf/icassp/PanayotovCPK15} is a collection of English speech of audiobook reading. Containing annotated texts for each utterance, Librispeech is an important corpus for speech recognition and text-to-speech synthesis tasks. However, Librispeech lacks varieties in terms of data source, language, and other speech aspects.

AliMeeting\cite{DBLP:conf/icassp/YuZFXZDHGYMXB22} is collected in a similar way as 3D-Speaker. Multiple recording devices are placed randomly in front of speakers during each recording session. However, the labels of devices and distance to speakers are not provided in AliMeeting. Containing fewer than 500 speakers, AliMeeting is not suitable to be used solely as a training corpus for speaker verification task.

The NIST SRE datasets are collected accumulated from the regularly held evaluations\cite{NISTSRE12}. However, it is not freely accessible to public.

There are many other audio datasets containing speaker identities, including but not limited to SITW\cite{DBLP:conf/interspeech/McLarenFCL16} with 300 speakers, AISHELL-4\cite{DBLP:conf/interspeech/FuCLJKCHXWBXDC21} with 61 speakers, and TIMIT\cite{TIMIT} with 630 speakers, etc.

\begin{table}[]
\caption{Comparison of several freely available audio datasets that provide speaker labels.}
\begin{tabular}{ccccccc}
\hline
              & \begin{tabular}[c]{@{}c@{}}\# of \\ Speakers\end{tabular} & \begin{tabular}[c]{@{}c@{}}Labels of \\ 
              Multiple \\
              Devices\end{tabular} & \begin{tabular}[c]{@{}c@{}}Labels of \\
              Multiple \\
              Distances\end{tabular} & \begin{tabular}[c]{@{}c@{}}Labels of \\
              Multiple \\
              Dialects\\ /Languages\end{tabular} & \begin{tabular}[c]{@{}c@{}}Sampling \\ Rate\end{tabular} & \begin{tabular}[c]{@{}c@{}}Has \\ annotated \\ texts\end{tabular} \\ \hline
\textbf{3D-Speaker}    & 10000+                                                        & Yes                                                          & Yes                                                            & Yes                                                                        & 16k \& 48k                                                                                                 & Yes                                                                \\ 
VoxCeleb 1\&2\cite{DBLP:conf/interspeech/NagraniCZ17} & 7000+                                                         & No                                                           & No                                                             & No                                                                         & 16k                                                                                                        & No                                                                \\ 
CN-Celeb\cite{DBLP:conf/icassp/FanKLLCCZZCW20}      & 3000                                                          & No                                                           & No                                                             & No                                                                         & 16k                                                                                                        & No                                                                \\ 
Librispeech\cite{DBLP:conf/icassp/PanayotovCPK15}   & 2497                                                          & No                                                           & No                                                             & No                                                                         & 16k                                                                                                         & Yes                                                               \\ 
AliMeeting\cite{DBLP:journals/corr/abs-2303-00332}    & 481                                                           & No                                                           & No                                                             & No                                                                         & 16k \& 48k                                                                                                  & Yes                                                               \\ 
AISHELL-4\cite{DBLP:conf/interspeech/FuCLJKCHXWBXDC21}     & 61                                                            & No                                                           & No                                                             & No                                                                         & 16k                                                                                                      & Yes                                                               \\ \hline
\end{tabular}
\label{TableCompare}
\end{table}

\section{Dataset Description}

\begin{figure}[h]
\centering
\includegraphics[width=\textwidth]{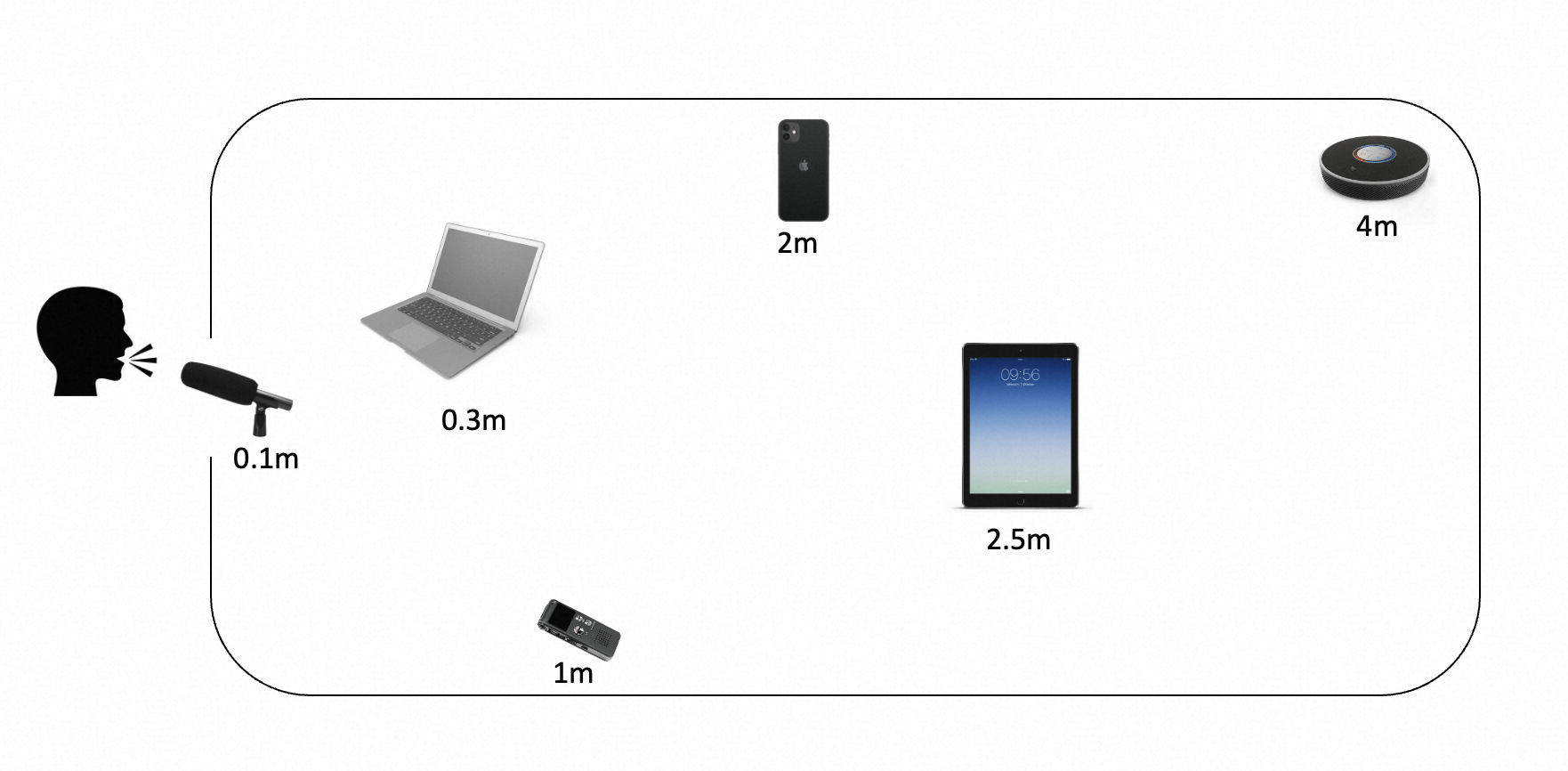}
\caption{An example of device placements in a recording session. Devices are shuffled randomly at the beginning of each recording session. }
\end{figure}

The training dataset includes a total of 10,000 speakers, and 579,013 utterances. The total duration of valid speech is $1124$ hours. It is worth noting that certain utterances in the dataset share identical speech content, as they were simultaneously recorded using different devices from varying distances. Additionally, the dataset features 1,200 speakers recorded speaking in two distinct dialects - standard Mandarin, and a regional dialect of the speakers' own choice.

\subsection{Multi-Device}

Every utterances are simultaneously recorded by several different devices, selected from Table \ref{TableDevice}: iPads, Android phones, iPhones, microphone arrays(Array for short), PC laptops, recording pens (RP for short), single directional microphones, phones(unspecified). 

The microphone arrays consist of 8 directional microphones. We follow the design of differential circular array described in \cite{DBLP:conf/interspeech/HuangF20}, which have previously been used in AliMeeting dataset\cite{DBLP:conf/icassp/YuZFXZDHGYMXB22} and speaker diarization system\cite{DBLP:conf/icassp/ZhengHWSFY21}.

\begin{table}[!ht]
    \centering
    \caption{Detailed information of devices in 3D-Speaker.}
    \begin{tabular}{|c|c|c|}
    \hline
        Device & \# of Utterance & Percentage \\ \hline
        iPad & 65151 & 11.25\% \\ \hline
        Android & 65208 & 11.26\% \\ \hline
        iPhone & 65194 & 11.26\% \\ \hline
        Array & 90058 & 15.55\% \\ \hline
        PC & 96262 & 16.63\% \\ \hline
        RP & 106611 & 18.41\% \\ \hline
        Directional & 57838 & 9.99\% \\ \hline
        Phone(unspecified) & 32691 & 5.65\% \\ \hline
        Total & 579013 & 100.00\% \\ \hline
    \end{tabular}
    \label{TableDevice}
\end{table}

\subsection{Multi-Distance}

During each recording session, different devices are randomly positioned at varying distances from the speaker. These specific distances are classified and presented in Table \ref{TableDistance}. Distances range from 0.1m to 4m. To simulate real-world usage scenarios, PC laptops are exclusively situated within 1 meter from the speakers, while directional microphones are placed no further than 0.3 meters from the speakers.

\begin{table}
    \centering
    \caption{Detailed information of source-to-device distances in 3D-Speaker.}
    \begin{tabular}{|c|c|c|c|}
    \hline
        Distance(m)  & \# of Utterances & Duration(h) & Duration Percentage  \\ \hline
        0.1 & 14772 & 26.94 & 2.40\%  \\ \hline
        0.2 & 7845 & 14.15 & 1.26\%  \\ \hline
        0.3 & 26141 & 57.03 & 5.07\%  \\ \hline
        0.8 & 4903 & 8.94 & 0.79\%  \\ \hline
        0.9 & 463 & 0.85 & 0.08\%  \\ \hline
        1 & 83968 & 180.86 & 16.08\%  \\ \hline
        1.2 & 690 & 1.18 & 0.10\%  \\ \hline
        1.5 & 3770 & 6.66 & 0.59\%  \\ \hline
        2 & 65596 & 138.00 & 12.27\%  \\ \hline
        2.5 & 4043 & 7.13 & 0.63\%  \\ \hline
        3 & 65870 & 138.48 & 12.31\%  \\ \hline
        4 & 115203 & 247.03 & 21.97\%  \\ \hline
        Unspecified & 185749 & 297.29 & 26.44\% \\ \hline
        Total & 579013 & 1124.52 & 100.00\% \\ \hline
    \end{tabular}
    \label{TableDistance}
\end{table}

\subsection{Multi-Dialect}

In training set we include 1074 speakers with multiple dialects, as illustrated in Table \ref{TableDialect}. Each of these speakers are first recorded speaking standard mandarin. Then they are asked to speak their own regional dialect. The entire session are recorded by multiple devices locating at different distances from the speaker. The selection of dialects was carried out with the aim of ensuring that they are significantly different from one another and from standard Mandarin. This is to the extent that individuals who do not speak the particular dialect would find it hardly comprehensible.

\begin{table}
\centering
    \caption{Detailed information of different dialects spoken in 3D-Speaker.}
    \begin{tabular}{|c|c|c|}
    \hline
        Dialect & \# of Speakers & Duration(h)  \\ \hline
        JiangHuai Mandarin & 30 & 2.027  \\ \hline
        Gan Dialect & 37 & 1.959  \\ \hline
        Wu Dialect & 130 & 8.442  \\ \hline
        Jin Dialect & 518 & 29.108  \\ \hline
        Min Dialect & 23 & 1.041  \\ \hline
        Central Plains Mandarin & 25 & 3.684  \\ \hline
        Hakka Dialect & 39 & 2.147  \\ \hline
        JiLu Mandarin & 12 & 5.518  \\ \hline
        LiaoJiao Mandarin & 13 & 6.443  \\ \hline
        Northern Mandarin & 2 & 0.934  \\ \hline
        Xiang Dialect & 2 & 0.106  \\ \hline
        Southwestern Mandarin & 238 & 13.44  \\ \hline
        Cantonese & 42 & 2.395  \\ \hline
        Total & 1074 & 77.244 \\ \hline
    \end{tabular}
    \label{TableDialect}
\end{table}

\subsection{Evaluation set}

Table \ref{TableEvalSet} provide descriptive information of evaluation set. There are a total of 240 speakers and 18782 utterances. None speakers are included in train set. The evaluation set includes 11 distinct dialects, all of which are spoken by some speakers in the train set.

\begin{table}[!ht]
    \centering
    \caption{Detailed information of evaluation set.}
    \begin{tabular}{|c|c|}
    \hline
        \# of Speakers & 240 \\ \hline
        \# of Utterances & 18782 \\ \hline
        \# of Dialects & 11 \\ \hline
        Duration(h) & 15.42 \\ \hline
    \end{tabular}

    \label{TableEvalSet}
\end{table}

\section{Experiments and Benchmarks}

The microphone array consists of 8 channels, each of which has a sampling rate of 48kHz. In our baseline systems, we only take the first channel and downsample it to 16kHz. In our previous studies we discovered that valuable information could be learned by modeling all 8 channels\cite{DBLP:conf/interspeech/ZhangZHLSFY21}\cite{DBLP:journals/corr/abs-2109-04049}.

For baseline systems, we choose CAM++ \cite{DBLP:journals/corr/abs-2303-00332}, ERes2Net\cite{DBLP:journals/corr/abs-2305-12838}, and ECAPA-TDNN\cite{DBLP:conf/interspeech/DesplanquesTD20}. The results are listed in Table \ref{TableResults}. We use EER and minDCF(p\_target=0.01,c\_miss=1,c\_fa=1) as the metrics for all experiments.

\subsection{Track A: Cross-Device Speaker Verification}

In cross-device trial, we guarantee that the enrollment and test utterances are recorded using separate devices. We also ensure that speech content differs between enrollment and test utterances for each of the 180,000 trials. The trial considers the ``iPhone'', ``Android'', and ``Phone'' categories as one, due to their acoustic similarities. 

\begin{table}[h!]
\centering
\caption{Performance of baseline systems on different tracks.}
\begin{tabular}{|c|c|cc|cc|cc|}
\hline
\multirow{2}{*}{Method} & \multirow{2}{*}{\# of Params} & \multicolumn{2}{c|}{Cross-Device}     & \multicolumn{2}{c|}{Cross-Distance}   & \multicolumn{2}{c|}{Cross-Dialect}    \\ \cline{3-8} 
                        &                                   & \multicolumn{1}{c|}{EER(\%)} & minDCF & \multicolumn{1}{c|}{EER(\%)} & minDCF & \multicolumn{1}{c|}{EER(\%)} & minDCF \\ \hline
                        ECAPA-TDNN\cite{DBLP:conf/interspeech/DesplanquesTD20}\tablefootnote{Implementation: \url{https://github.com/speechbrain/speechbrain/blob/develop/speechbrain/lobes/models/ECAPA_TDNN.py}} &     20.8M                              & \multicolumn{1}{c|}{8.87}        &   0.732     & \multicolumn{1}{c|}{12.26}        &    0.805    & \multicolumn{1}{c|}{14.53}        &  0.913      \\ \hline
                        CAM++ Base\cite{DBLP:journals/corr/abs-2303-00332}\tablefootnote{Official implementation: \url{https://github.com/alibaba-damo-academy/3D-Speaker/tree/main/egs/sv-cam\%2B\%2B}} &     7.2M                              & \multicolumn{1}{c|}{7.75}        &  0.723      & \multicolumn{1}{c|}{11.29}        &  0.783      & \multicolumn{1}{c|}{13.44}        &    0.886    \\ \hline
                        ERes2Net Base\cite{DBLP:journals/corr/abs-2305-12838} \tablefootnote{Official implementation: \url{https://github.com/alibaba-damo-academy/3D-Speaker/tree/main/egs/sv-eres2net}}&    4.6M                              & \multicolumn{1}{c|}{7.06}        &    0.656    & \multicolumn{1}{c|}{9.95}        &   0.753     & \multicolumn{1}{c|}{12.76}        &     0.871  \\ \hline                        
                        ERes2Net Large\cite{DBLP:journals/corr/abs-2305-12838}\tablefootnote{Official implementation: \url{https://github.com/alibaba-damo-academy/3D-Speaker/tree/main/egs/sv-eres2net}} &    18.3M                              & \multicolumn{1}{c|}{6.55}        &     0.640   & \multicolumn{1}{c|}{9.45}        &  0.713      & \multicolumn{1}{c|}{11.01}        &   0.811     \\ \hline
\end{tabular}
\label{TableResults}
\end{table}

\subsection{Track B: Cross-Distance Speaker Verification}

In cross-distance trial, a distance of 0.8 meters or greater is considered ``far-field'', while distances less than 0.8 meters are classified as ``near-field''. With this categorization in mind, we meticulously ensure that, across all 175,163 trials, the enrollment and test utterances are selected from different classification categories. Similar to Track A, we guarantee that the speech content of the enrollment and test utterances differs from one another.

\subsection{Track C: Cross-Dialect Speaker Verification}

In cross-dialect trial, it is guaranteed that either the enrollment or test utterance is standard mandarin, while the other is the regional dialect of the corresponding speaker. 

\subsection{Track D: Language/Dialect Identification}

In language/dialect identification task, we use all utterances in the test set and estimate the overall identification accuracy. We provide a baseline benchmark using vanilla CAM++. To overcome imbalance of labels, we only use a small subset of training data in the baseline system. The results are listed in Table \ref{dialectres}.

\begin{table}[]
\centering
\caption{Performance of baseline system on dialect identification.}
\begin{tabular}{|c|c|c|}
\hline
                & Train Accuracy(\%) & Test Accuracy(\%) \\ \hline
Baseline\cite{DBLP:journals/corr/abs-2303-00332} & 96.82                 & 29.36             \\ \hline
\end{tabular}
\label{dialectres}
\end{table}

\subsection{Other tasks}

Other than the tasks and benchmarks described above, the rich multi-domain information in 3D-Speaker allows researchers to design tasks of their own and tailor training and evaluation set to meet their needs.

\textbf{Out-of-domain learning.} 3D-Speaker allows researchers to carry out experiments on out-of-domain learning. For example, researchers could remove utterances from certain devices from training set and evaluate the model performance on these devices. One could also train the model only on ``near-field'' data and evaluate them on ``far-field'' data.

\textbf{Self-supervised learning.} The diverse nature of 3D-Speaker makes it an ideal candidate for exploring self-supervised learning methods on acoustic data. In Table \ref{RDINO} we provide a baseline system using RDINO self-supervised learning method, in which we treat all labels in 3D-Speaker as unknown\cite{DBLP:journals/corr/abs-2211-04168}.

\begin{table}[]
\centering
\caption{Performance of baseline self-supervised learning system on different tracks. EER and minDCF$( p_{target}=0.05,c_{miss}=1,c_{fa}=1)$ are used to measure the performance.}
\begin{tabular}{|c|cc|cc|cc|}
\hline
\multirow{2}{*}{Method} & \multicolumn{2}{c|}{Cross-Device}     & \multicolumn{2}{c|}{Cross-Distance}   & \multicolumn{2}{c|}{Cross-Dialect}    \\ \cline{2-7} 
                        &                                    \multicolumn{1}{c|}{EER(\%)} & minDCF & \multicolumn{1}{c|}{EER(\%)} & minDCF & \multicolumn{1}{c|}{EER(\%)} & minDCF \\ \hline
                        
                        RDINO\cite{DBLP:journals/corr/abs-2211-04168}\tablefootnote{Official implementation: \url{https://github.com/alibaba-damo-academy/3D-Speaker/tree/main/egs/sv-rdino}}
                        &                                   \multicolumn{1}{c|}{20.41}       &   0.972     & \multicolumn{1}{c|}{21.92}        &   0.966     &  \multicolumn{1}{c|}{25.53}        &  0.999      \\ \hline

\end{tabular}
\label{RDINO}
\end{table}

\textbf{Evaluate large universal speech models.} 3D-Speaker is a suitable resource to evaluate the universal performance of large speech models. Large universal speech models are expected to perform reasonably well on various domains.

\section{Conclusion}

We introduced 3D-Speaker, a large-scale speech corpus designed to facilitate the research of speech representation disentanglement. The controlled combinations of multi-dimensional audio data in this corpus yield a matrix of a diverse blend of speech representation entanglement, motivating intriguing methods to untangle them. The multi-domain nature of 3D-Speaker also makes it a suitable resource to evaluate large universal speech models and experiment methods of out-of-domain learning and self-supervised learning. Additionally, 3D-Speaker is the largest publicly-accessible corpus in terms of number of speakers, which can be used to improve the performance of speaker verification systems and other speech-related tasks. Overall, 3D-Speaker provides a valuable resource for advancing the research of speech-related fields.

\section{Ethics}

We understand that voice is a unique physical characteristics and important human biometrics. Therefore, during the collection of 3D-Speaker, we ensure that mutual agreements are reached with the speakers. The speakers understand that the recorded content will be used for the purpose of academic research and be made publicly accessible.

\bibliography{refs.bib}
\end{document}